\documentclass[letterpaper]{article}
\usepackage{aaai}
\usepackage{rotating}
\usepackage{times}
\usepackage{latexsym}
\usepackage{epic}
\usepackage{amsfonts}
\usepackage{amsmath}
\usepackage{amssymb}
\usepackage{amsthm}
\usepackage{xspace}
\usepackage{url}
\usepackage{epstopdf}

\ifx\pdfoutput\undefined  \else
 \fi

\usepackage[ruled,linesnumbered,boxed]{algorithm2e}

\input{epsf}
\newtheorem{definition}{Definition}
\newtheorem{myexample}{Example}

\newtheorem{corollary}{Corollary}

\newcommand{\myOmit}[1]{}

\newcommand{\mf}{{\bf\mathcal F}}
\newcommand{\mi}{{\bf\mathcal I}}
\newcommand{\mc}{\mathcal C}
\newcommand{\ra}{\rightarrow}
\newcommand{\alt}{{\rm Alt}}
\newcommand{\rev}{{\rm Rev}}
\newcommand{\others}{{\rm Others}}
\newcommand{\borda}{{\rm Borda}}
\newcommand{\wmg}{{\rm WMG}}
\newcommand{\mv}{{\mathcal V}}
\newcommand{\ms}{{\mathcal S}}
\newcommand{\mr}{{\mathcal R}}

\newtheorem{thm}{Theorem}

\newtheorem{prop}{Proposition}

\renewenvironment{proof}{\noindent {\bf Proof.}}{\hfill$\Box$}

\begin{document}
\title{Dominating Manipulations in Voting with Partial Information}
\author{Vincent Conitzer\\ Department of Computer Science\\ Duke University \\ Durham, NC 27708, USA\\
conitzer@cs.duke.edu\And Toby Walsh\\
NICTA and UNSW\\ Sydney, Australia\\
toby.walsh@nicta.com.au
 \And Lirong Xia\\ Department of Computer Science\\ Duke University \\ Durham, NC 27708, USA\\
lxia@cs.duke.edu}
 \maketitle
\begin{abstract}
We consider manipulation problems when the
manipulator only has partial information about the votes
of the non-manipulators. Such partial information is
described by an {\em information
set}, which is the set of profiles of the non-manipulators that are
indistinguishable to the manipulator. Given such an information set, a
{\em dominating manipulation} is a non-truthful vote that the manipulator
can cast which makes the winner at least as preferable (and sometimes
more preferable) as the winner when the manipulator votes truthfully.
When the manipulator has full information,
computing whether or not there exists a dominating
manipulation is in {\sf P} for many common voting rules
(by known results). We show that when the
manipulator has no information,
there is no dominating manipulation
for many common voting rules. When the manipulator's information
is represented by partial orders and only a small portion of the
preferences are unknown, computing a
dominating manipulation is {\sf NP}-hard
for many common voting rules. Our results
thus throw light on whether we can prevent strategic behavior
by limiting information about the votes of other voters.
\end{abstract}

\section{Introduction}
In computational social choice, one appealing escape from the
Gibbard-Satterthwaite
theorem~\cite{Gibbard73:Manipulation,Satterthwaite75:Strategy} was
proposed in~\cite{Bartholdi89:Computational}. Whilst manipulation
may always be possible, perhaps it is computationally too difficult
to find? Many results have subsequently been proven showing that
various voting rules are {\sf NP}-hard to
manipulate~\cite{Bartholdi91:Single,Conitzer03:Universal,Elkind05:Hybrid,Conitzer07:When,Faliszewski08:Copeland,Xia09:Complexity,Faliszewski10:Manipulation}
in various senses. However, recent results suggest that computing a
manipulation is easy on average or in many cases. Therefore,
computational complexity seems to be a weak barrier against
manipulation. See~\cite{Faliszewski10:Using,Faliszewski10:AI} for
some surveys of this recent research.

It is normally assumed that the manipulator has
full information about the votes of the non-manipulators. The
argument often given is that if it is NP-hard with full
information, then it only can be at least as computationally
difficult with partial information.
%Note that this argument does not reverse. If it is computationally
%easy to compute a manipulation with full information, then we cannot
%say how hard it will be with partial information.
However, when there is only one manipulator, computing a
manipulation is polynomial for most common voting rules, including all
positional scoring rules, Copeland, maximin, and voting trees. The
only known exceptions are STV~\cite{Bartholdi91:Single} and ranked
pairs~\cite{Xia09:Complexity}. Therefore, it is not clear whether
a single manipulator has incentive to lie when the manipulator
only has partial information.
%The previous results about
%manipulation being polynomial to compute for many voting rules
%with complete information tell us nothing about the
%computational complexity of computing a manipulation
%with partial information.

In this paper, we study the problem of how one manipulator
computes a manipulation based on partial information about
the other votes.
%We argue that it may be reasonable to assume that the
%manipulator has some partial information about the votes of the
%non-manipulators.
For example, the manipulator may know that some voters prefer
one alternative to another, but might not be able to know
all pairwise comparisons for all voters. We suppose
the knowledge of the manipulator is described by an {\em information set} $E$.
This is some subset of possible profiles of the non-manipulators
which is known to contain the true profile.
Given an information set and a pair of votes $U$ and $V$,
if for every profile in $E$, the manipulator is not worse off voting $U$ than voting $V$,
and there exists a profile in $E$ such that the manipulator is strictly better
off voting $U$, then we say that $U$ {\em %(weakly)
dominates} $V$. If there exists a vote $U$ that dominates the true
preferences of the manipulator then the manipulator has an incentive
to vote untruthfully. We call this a {\em dominating manipulation.}
If there is no such vote, then a risk-averse manipulator might have
little incentive to vote strategically.

We are interested in whether a voting rule $r$ is {\em immune} to
dominating manipulations, meaning that a voter's true preferences
are never dominated by another vote. If $r$ is not immune to
dominating manipulations, we are interested in whether $r$ is {\em
resistant}, meaning that computing whether a voter's true
preferences are dominated by another vote $U$ is {\sf NP}-hard, or
{\em vulnerable}, meaning that this problem is in {\sf P}. These
properties depend on both the voting rule and the form of the
partial information. Interestingly, it is not hard to see that most
voting rules are immune to manipulation when the partial information
is just the current winner. For instance, with any majority
consistent rule (for example, plurality), a risk averse manipulator
will still want to vote for her most preferred alternative. This
means that the chairman does not need to keep the current winner
secret to prevent such manipulations. On the other hand, if the
chairman lets slip more information, many rules stop being immune.
With most scoring rules, if the manipulator knows the current
scores, then the rule is no longer immune to such manipulation. For
instance, when her most preferred alternative is too far behind to
win, the manipulator might vote instead for a less preferred
candidate who can win.

In this paper, we focus on the case where the partial information is
represented by a profile $P_{po}$ of partial orders, and the
information set $E$ consists of all linear orders that extend
$P_{po}$. The dominating manipulation problem is related to the {\em
possible/necessary winner}
problems~\cite{Konczak05:Voting,Walsh07:Uncertainty,Betzler09:Multivariate,Betzler10:Towards,Xia11:Determining}.
In possible/necessary winner problems, we are given an alternative
$c$ and a profile of partial orders $P_{po}$ that represents the
partial information of the voters' preferences. We are asked whether
$c$ is the winner for {\em some} extension of $P_{po}$ (that is, $c$
is a {\em possible winner}), or whether $c$ is the winner for {\em
every} extension of $P_{po}$ (that is, $c$ is a {\em necessary
winner}). We note that in the possible/necessary winner problems,
there is no manipulator and $P_{po}$ represents the chair's partial
information  about the votes. In dominating manipulation problems,
$P_{po}$ represents the partial information of the manipulator about
the non-manipulators.
%Moreover, we are not directly asked which candidate will be
%the winner. We are asked whether the manipulator has an incentive to
%cast a false vote such that the winner is always as good as (and
%sometimes better than) the winner as if she cast a truthful vote,
%for all extensions of $P_{po}$.

We start with the special case where the manipulator has complete
information. In this setting the dominating manipulation problem
reduces to the standard manipulation problem, and many common voting
rules are vulnerable to dominating manipulation (from known
results). When the manipulator has no information, we show that a
wide range of common voting rules are immune to dominating
manipulation.
%, including all Condorcet
%consistent rules, Borda, all positional scoring rule (when the
%number of non-manipulators is at least $6(m-2)$, where $m$ is the
%number alternatives.)
When the manipulator's partial information is
represented by partial orders, our results are summarized in
Table~\ref{tab:po}.
\begin{table}[htp]\vspace{-2mm}
\centering
\begin{tabular}{|r|c|c|}
\hline & {\sc dominating manipulation\footnotemark[1]}\\
\hline STV& Resistant (Proposition~\ref{prop:harddm})\\
\hline Ranked pairs& Resistant (Proposition~\ref{prop:harddm})\\
\hline Borda& Resistant (Theorem~\ref{thm:hardborda})\\
\hline Copeland& Resistant (Corollary~\ref{coro:WMGDMhard})\\
\hline Voting trees& Resistant (Corollary~\ref{coro:WMGDMhard})\\
\hline Maximin& Resistant (Theorem~\ref{thm:DMmaximin})\\
\hline Plurality& Vulnerable (Algorithm~\ref{alg:domination})\\
\hline Veto& Vulnerable \begin{tabular}{c}(Omitted due to\\ the space constraint.)\end{tabular}\\
\hline
\end{tabular}
\caption{\footnotesize Computational complexity of the {dominating
manipulation} problems with partial orders, for common voting
rules.} \label{tab:po}\vspace{-2mm}
\end{table}
\footnotetext[1]{All hardness results hold even when the number of
undetermined pairs in each partial order is no more than a
constant.} \addtocounter{footnote}{1}

Our results are encouraging. For most voting rules $r$ we study in
this paper (except plurality and veto), hiding even a little
information makes $r$ resistant to dominating manipulation. If we
hide all information, then $r$ is immune to dominating manipulation.
Therefore, limiting the information available to the manipulator
appears to be a promising way to prevent strategic voting.
\vspace{-1mm}
\section{Preliminaries}
Let $\mc=\{c_1,\ldots,c_m\}$ be the set of {\em alternatives} (or
{\em
  candidates}).  A linear order on $\mc$ is a transitive, antisymmetric,
and total relation on $\mc$.  The set of all linear orders on $\mc$
is denoted by $L(\mc)$.  An $n$-voter profile $P$ on $\mc$ consists
of $n$ linear orders on $\mc$.  That is, $P=(V_1,\ldots,V_n)$, where
for every $j\leq n$, $V_j\in L(\mc)$. The set of all $n$-profiles is
denoted by $\mf_n$. We let $m$ denote the number of alternatives.
For any linear order $V\in L(\mc)$ and any $i\leq m$, $\alt(V,i)$ is
the alternative that is ranked in the $i$th position in $V$. A {\em
voting rule} $r$ is a function that maps any profile on $\mc$ to a
unique winning alternative, that is, $r:\mf_1\cup\mf_2\cup\ldots\ra
\mc$.
%a {\em voting correspondence} $c$ can
%select more than one winner, that is, $c:P(\mc)\ra 2^\mc$.
The following are some common voting rules. In this paper, if not
mentioned specifically, ties are broken in the fixed order $c_1\succ
c_2\succ\cdots\succ c_m$.

$\bullet$ {\em (Positional) scoring rules}: Given a {\em scoring
vector}
  $\vec s_m=(\vec s_m(1),\ldots,\vec s_m(m))$ of $m$ integers, for any vote $V\in L(\mc)$ and any $c\in
  \mc$, let $\vec s_m(V,c)=\vec s_m(j)$, where $j$ is the rank of $c$ in $V$.  For any
  profile $P=(V_1,\ldots,V_n)$, let $\vec s_m(P,c)=\sum_{j=1}^n\vec s_m(V_j,c)$.
  The rule will select $c\in \mc$ so that $\vec s_m(P,c)$ is
  maximized.
We assume %$\vec s_m(i)\in \mathbb N$ and $\vec s_m(1)>\vec s_m(m)$
%for any $i\leq m$,.
scores are integers and decreasing.
  Some examples of positional scoring rules are {\em Borda}, for which the scoring
  vector is $(m-1, m-2, \ldots,0)$, {\em plurality}, for which the
scoring vector is $(1,0,\ldots,0)$, and {\em veto}, for which the
scoring vector is $(1,\ldots,1,0)$.
%, and $k$-approval ($1\leq k\leq
%m-1$), for which the scoring vector is $(\underbrace{1,
%\ldots,1}_{k},0, \ldots,0)$.
%In this paper, we assume that

$\bullet$ {\em Copeland}: For any two alternatives $c_i$ and $c_j$,
we conduct a {\em pairwise election} in which we count how many votes
rank $c_i$ ahead of $c_j$, and how many rank $c_j$ ahead of $c_i$.
$c_i$ wins if and only if the majority of voters rank $c_i$ ahead of
$c_j$. An alternative receives one point for each such win in a
pairwise election. Typically, an alternative also receives half a
point for each pairwise tie, but this will not matter for our
results. The winner is the alternative with the highest score.

$\bullet$ {\em Maximin}: Let $D_P(c_i,c_j)$ be the number of
votes that
  rank $c_i$ ahead of $c_j$ minus the number of votes that
  rank $c_j$ ahead of $c_i$ in the profile $P$.  The winner is the alternative $c$ that
  maximizes $\min\{D_P(c,c'):c'\in \mc, c'\neq c\}$.
%\item {\em Bucklin}: An alternative $c$'s Bucklin score is the smallest
%  number $k$ such that more than half of the votes rank $c$ among the top
%  $k$ alternatives.  The winner is the alternative who has the smallest
%  Bucklin score.  (Sometimes, ties are broken by the number of votes that
%  rank an alternative among the top $k$, but for simplicity we will not
%  consider this tiebreaking rule here.)

$\bullet$  {\em Ranked pairs}: This rule first creates an entire
ranking of
  all the alternatives. % $D_P(c_i,c_j)$ is defined as for the maximin rule.
  In each step, we will consider a pair of alternatives $c_i, c_j$ that we
  have not previously considered; specifically, we choose the remaining
  pair with the highest $D_P(c_i,c_j)$.  We then fix the order $c_i\succ c_j$,
  unless this contradicts previous orders that we fixed (that is, it
  violates transitivity).  We continue until we have considered all pairs
  of alternatives (hence we have a full ranking).  The alternative at the
  top of the ranking wins.

$\bullet$  {\em Voting trees}: A voting tree is a binary tree with
$m$
  leaves, where each leaf is associated with an alternative. In each round,
  there is a pairwise election between an alternative $c_i$ and its sibling
  $c_j$: if the majority of voters prefer $c_i$ to $c_j$, then $c_j$ is
  eliminated, and $c_i$ is associated with the parent of these two nodes.
%  similarly, if the majority of voters prefer $c_j$ to $c_i$, then $c_i$ is
%  eliminated, and $c_j$ is associated with the parent of these two
%  nodes.
The alternative that is associated with the root of the tree (i.e. wins
  all its rounds) is the winner.
%\item[7.] {\em Plurality with runoff}: The rule has two steps. In
%the first step, all alternatives except the two that are ranked in
%the top position for most times are eliminated, and the votes
%transfers to the second round, in which the plurality rule (a.k.a.
%{\em majority} rule in case of two alternatives) is used to select
%the winner.

$\bullet$   {\em Single transferable vote (STV)}: The election has
$m$ rounds. In each round, the
  alternative that gets the lowest plurality score (the number of times that the alternative is ranked in the top position) drops out, and is removed
  from all of the votes (so that votes for this alternative transfer to
  another alternative in the next round). The last-remaining alternative is the
  winner.

For any profile $P$, we let $\wmg(P)$ denote the {\em weighted
majority graph} of $P$, defined as follows. $\wmg(P)$ is a directed
graph whose vertices are the alternatives. For $i \neq j$, if
$D_P(c_i,c_j) > 0$, then there is an edge $(c_i,c_j)$ with weight
$w_{ij} = D_P(c_i,c_j)$.
%Also, for $i < j$, if $D_P(c_i,c_j) = 0$,
%then there is an edge $(c_i,c_j)$ with weight $w_{ij}=0$.

We say that a voting rule $r$ is based on the {\em weighted majority
graph (WMG)}, if for any pair of profiles $P_1,P_2$ such that
$\wmg(P_1)=\wmg(P_2)$, we have $r(P_1)=r(P_2)$. A voting rule $r$ is
{\em Condorcet consistent} if it always selects the Condorcet winner
(that is, the alternative that wins each of its pairwise elections)
whenever one exists.
%For example, Copeland, maximin, ranked pairs,
%and voting trees are Condorcet consistent.

\vspace{-1mm}
\section{Manipulation with Partial Information}\vspace{-1mm}
We now introduce the framework of this paper. Suppose there are
$n\geq 1$ non-manipulators and one manipulator. The information the
manipulator has about the votes of the non-manipulators is
represented by an
%To better present the partial information that the manipulator
%obtains, we model the situation as a two-stage extensive-form game
%of imperfect information. In the first stage, the non-manipulators
%cast their votes, and in the second stage, the manipulators cast her
%vote. The manipulator receives a {\em signal} from a set of signals
%$\{e_1,\ldots,e_t\}$. For each $l\leq t$, $e_l$ corresponds to a
%subset of profiles $E_l\subseteq \mf_n$, called an
{\em information set} $E$. The manipulator knows for sure that the
profile of the non-manipulators is in $E$. However, the manipulator
does not know exactly which profile in $E$ it is. Usually $E$ is
represented in a compact way. Let $\mi$ denote the set of all
possible information sets in which the manipulator may find herself.
%Figure~\ref{fig:game} for an illustration.
%\begin{figure}[htp]
%\centering \includegraphics[width=9cm]{fig3.eps}\vspace{-2mm}
%\caption{\footnotesize The manipulator's information sets, where
%$K=(m!)^n$.}\vspace{-1mm} \label{fig:game}
%\end{figure}

\begin{myexample}
Suppose the voting rule is $r$.

$\bullet$ If the manipulator has no information, then the only
information set is $E=\mf_n$. Therefore $\mi=\{\mf_n\}$.

$\bullet$ If the manipulator has complete information, then
$\mi=\{\{P\}: P\in \mf_n\}$.

$\bullet$ If the manipulator knows the current winner (before the
manipulator votes), then the set of all information sets the
manipulator might know is $\mi=\{E_1,E_2,\ldots,E_m\}$, where for
any $i\leq m$, $E_i=\{P\in\mf_n:r(P)=c_i\}$.
\end{myexample}
%Now, given an information set $E$, a manipulator's strategy is a
%mapping that assigns each signal to a vote she cast.
Let $V_M$ denote the true preferences of the manipulator. Given a voting
rule $r$ and an information set $E$, we say that a vote $U$ {\em
%(weakly)
dominates} another vote $V$, if for every profile $P\in E$, we have
$r(P\cup\{U\})\succeq_{V_M} r(P\cup \{V\})$, and there exists $P'\in
E$ such that $r(P'\cup\{U\})\succ_{V_M} r(P'\cup \{V\})$. In other
words, when the manipulator only knows the voting rule $r$ and the
fact that the profile of the non-manipulators is in $E$ (and no
other information), voting $U$ is a strategy that
%weakly
dominates voting $V$.
%any profile in $E$, she is not worse off, and in at least one
%profile in $E$ she is better off, compared to voting her true
%preferences. Formally, Given an information set $E$, a voting rule
%$r$, the true preferences $V$ of the manipulators, and two votes $U$
%and $W$, we say $W$ {\em (weakly) dominates} $U$, if for any $P\in
%E$, $r(P\cup \{W\})\succeq_V r(P\cup \{U\})$ and there exists $P'\in
%E$ such that $r(P\cup \{U\})\succ_V r(P\cup \{U\})$.
We define the following two decision problems.
\begin{definition}
Given a voting rule $r$, an information set $E$, the true
preferences $V_M$ of the manipulator, and two votes $V$ and $U$, we
are asked the following two questions.\\
$\bullet$ Does $U$ dominate $V$? This is the
{\sc domination} problem.\\
$\bullet$ Does there exist a vote $V'$ that dominates $V_M$? This is
the {\sc dominating manipulation} problem.
\end{definition}
We stress that usually $E$ is represented in a compact way,
otherwise the input size would already be exponentially large, which
would trivialize the computational problems. Given a set $\mi$ of
information sets, we say a voting rule $r$ is {\em immune} to
dominating manipulation, if for every $E\in \mi$ and every $V_M$
that represents the manipulator's preferences, $V_M$ is not
dominated; $r$ is {\em resistant} to dominating manipulation, if
{\sc dominating manipulation} is {\sf NP}-hard (which means that $r$
is not immune to dominating manipulation, assuming {\sf P}$\neq${\sf
NP}); and $r$ is {\em vulnerable} to dominating manipulation, if $r$
is not immune to dominating manipulation, and {\sc dominating
manipulation} is in {\sf P}.

\section{Manipulation with Complete/No Information}
In this section we focus on  the following two special cases: (1)
the manipulator has complete information, and (2) the manipulator has no
information.
%The formal case corresponds to $\mi=\{\{P\}:P\in
%\mf_{n}\}$, and the latter case corresponds to $\mi=\{\mf_{n}\}$.
%
It is not hard to see that when the manipulator has complete
information, {\sc dominating manipulation} coincides with the
standard manipulation problem. Therefore, our framework of
dominating manipulation is an extension of the traditional
manipulation problem, and we immediately obtain the following
proposition from the Gibbard-Satterthwaite
theorem~\cite{Gibbard73:Manipulation,Satterthwaite75:Strategy}.
\vspace{-1mm}
\begin{prop}
\label{prop:GS}  When $m\geq 3$ and the manipulator has full
information, a voting rule satisfies non-imposition and is immune to
dominating manipulation if and only if it is a dictatorship.
\end{prop}\vspace{-1mm}

%It is easy to see that for any voting rule $r$ under which computing
%the winner is in {\sf P}, {\sc domination} is in {\sf P}; and if the
%coalitional manipulation problem is {\sf NP}-complete (respectively,
%{\sf P}), then {\sc dominating manipulation} is in {\sf NP}-complete
%(respectively, {\sf P}). Therefore,
The following proposition directly follows from the computational
complexity of the manipulation problems for some common voting
rules~\cite{Bartholdi89:Computational,Bartholdi91:Single,Conitzer07:When,Zuckerman09:Algorithms,Xia09:Complexity}.
\vspace{-1mm}\begin{prop} \label{prop:harddm} When the manipulator
has complete information, STV and ranked pairs are resistant to {\sc
dominating manipulation}; all positional scoring rules, Copeland,
voting trees, and maximin are vulnerable to dominating manipulation.
\end{prop}\vspace{-1mm}

Next, we investigate the case where the manipulator has no
information. We obtain the following positive results. Due to the
space constraint, most proofs are omitted.
%a
%preliminary full version can be found via the
%following anonymous link:\\
%\url{http://www.4shared.com/document/ooIzocgC}
\vspace{-1mm}
\begin{thm}
\label{thm:ccnoinfo} When the manipulator has no information, any
Condorcet consistent voting rule $r$ is immune to dominating
manipulation.
\end{thm}
\myOmit{\begin{proof} For the sake of contradiction, let $U$
dominates $V_M$. Because $U\neq V_M$, there exist two alternatives
$a$ and $b$ such that $a\succ_{V_M} b$ and $b\succ_U a$. We prove
the theorem in the following two cases.

{\bf Case 1:} $n$ is even. For any $j$ such that $1\leq j\leq n/2$,
we let $V_{2j-1}=[a\succ b\succ (\mc\setminus\{a,b\})]$, where the
alternatives in $\mc\setminus\{a,b\}$ are ranked according to the
ascending order of their subscripts; let $V_{2j}=[b\succ a\succ
\rev(\mc\setminus\{a,b\})]$. Here $ \rev(\mc\setminus\{a,b\})$ is
the reverse of $\mc\setminus\{a,b\}$. Let $P=(V_1,\ldots,V_{n})$. It
follows that $a$ is the Condorcet winner for $P\cup\{V_M\}$ and $b$
is the Condorcet winner for $P\cup\{U\}$. Because $a\succ_{V_M} b$,
$V_M$ is not dominated by $U$, which contradicts the assumption.

{\bf Case 2:} $n$ is odd. For any $j$ such that $1\leq j\leq
(n-1)/2$, we let $V_{2j-1}=[a\succ b\succ (\mc\setminus\{a,b\})]$
and $V_{2j}=[b\succ a\succ \rev(\mc\setminus\{a,b\})]$. Suppose
$a=c_{i_1}$ and $b=c_{i_2}$. Let $V_{n}=\left\{\begin{array}{ll}
V_1&\text{if }i_1>i_2\\
V_2&\text{if }i_1<i_2\end{array}\right.$. Let
$P=(V_1,\ldots,V_{n})$. It follows that $a$ is the Condorcet winner
for $P\cup\{V_M\}$ and $b$ is the Condorcet winner for $P\cup\{U\}$,
which contradicts the assumption.
\end{proof}
} \vspace{-1mm}
\begin{thm}
\label{thm:bordanoinfo} When the manipulator has no information,
Borda is immune to dominating manipulation.
\end{thm}\vspace{-1mm}
\vspace{-1mm}
\begin{thm}
\label{thm:psnoinfo}When the manipulator has no information and
$n\geq 6(m-2)$, any positional scoring rule is immune to dominating
manipulation.
\end{thm}

These results demonstrate that the information that the manipulator
has about the votes of the non-manipulators plays an important role
in determining strategic behavior. When the manipulator has complete
information, many common voting rules are vulnerable to dominating
manipulation, but if the manipulator has no information, then many
common voting rules become immune to dominating manipulation.
\vspace{-2mm}
\section{Manipulation with Partial Orders}\vspace{-1mm}
In this section, we study the case where the manipulator has partial
information about the votes of the non-manipulators. We suppose the
information is represented by a profile $P_{po}$ composed of partial
orders. That is, the information set is $E=\{P\in\mf_n: P\text{
extends }P_{po}\}$. We note that the two cases discussed in the
previous section (complete information and no information) are
special cases of manipulation with partial orders. Consequently, by
Proposition~\ref{prop:GS}, when the manipulator's information is
represented by partial orders and $m\geq 3$, no voting rule that
satisfies non-imposition and non-dictatorship is immune to
dominating manipulation. It also follows from
Theorem~\ref{thm:bordanoinfo} that STV and ranked pairs are
resistant to dominating manipulation.
%We define the {\sc domination} with partial order problem as
%follows. Given a voting rule $r$, a profile of partial orders
%$P_{po}$, the true preferences $V$ of the manipulator, and two votes
%$U$ and $W$, we are asked whether $W$ dominates $U$ under $r$ when
%$E=\{P: P\text{ is a profile of linear orders and is an extension of
%}P_{po} \}$. Similarly we define the {\sc dominating manipulation}
%problem with partial orders.
The next theorem states that even when the manipulator only misses a
tiny portion of the information, Borda becomes resistant to
dominating manipulation.\vspace{-1.5mm}
\begin{thm}
\label{thm:hardborda} {\sc domination} and {\sc dominating
manipulation} with partial orders are {\sf NP}-hard for Borda, even
when the number of unknown pairs in each vote is no more than $4$.
\end{thm}\vspace{-1.5mm}
\begin{proof} We only prove that {\sc domination} is {\sf NP}-hard, via a
reduction from {\sc Exact Cover by 3-Sets (x3c)}. The proof for {\sc
dominating manipulation} is omitted due to space constraint. The
reduction is similar to the proof of the {\sf NP}-hardness of the
possible winner problems under positional scoring rules
in~\cite{Xia11:Determining}.
%However, we were unable to find a
%general connection between the complexity of the {\sc domination}
%problem and the complexity of the possible winner problem.

In an {\sc x3c} instance, we are given two sets
$\mv=\{v_1,\ldots,v_q\}$, $\ms=\{S_1,\ldots,S_t\}$, where for any
$j\leq t$, $S_j\subseteq \mv$ and $|S_j|=3$. We are asked whether
there exists a subset $\ms'$ of $\ms$ such that each element in
$\mv$ is in exactly one of the 3-sets in $\ms'$. We construct a {\sc domination} instance as follows.\\
{\bf Alternatives:} $\mc=\{c,w,d\}\cup \mv$, where $d$ is an
auxiliary alternative. Therefore, $m=|\mc|=q+3$. Ties are broken in the following order: $c\succ w\succ \mv\succ d$. \\
{\bf Manipulator's preferences and possible manipulation:}
$V_M=[w\succ c\succ d\succ \mv]$. We are asked whether $V=V_M$ is
dominated by $U=[w\succ d\succ c\succ\mv]$.

\noindent{\bf The profile of partial orders:} Let $P_{po}=P_1\cup
P_2$, defined as follows.

\noindent{\bf First part ($P_1$) of the profile:} For each $j\leq
t$,
% choose an arbitrary set $B_j\subset \mc\setminus (S_j\cup
%\{w,d\})$ with $|B_j|=k-1$.
%Let $O(B_i,w,S_i,d,Others)$ be some
%linear order that agrees with $B_i\succ w\succ S_i\succ d\succ
%Others$.
We define a partial order $O_{j}$ as follows.\\
$\hfill O_{j}=[w\succ S_j\succ d\succ
\text{Others}]\setminus[\{w\}\times (S_j\cup\{d\})]\hfill$

That is, $O_{j}$ is a partial order that agrees with $w\succ
S_j\succ d\succ \text{Others}$, except that the pairwise relations
between $(w,S_j)$ and $(w,d)$ are not determined (and these are the
only $4$ unknown relations).
%In the definition,
%the sets, e.g.~$B_i$, $S_i$, $Others$, are converted to linear
%orders in an obvious way. The way to convert them does not affect
%the proof because we have the second part to keep balance.
Let $P_1=\{O_{1},\ldots,O_{t}\}$.\\
{\bf Second part ($P_2$) of the profile:} We first give the
properties that we need $P_2$ to satisfy, then show how to construct
$P_2$ in polynomial time. All votes in $P_2$ are linear orders that
are used to adjust the score differences between alternatives. Let
$P_1'=\{w\succ S_i\succ d\succ \text{Others}: i\leq t\}$. That is,
$P_1'$ ($|P_1'| = t$) is an extension of $P_1$ (in fact, $P_1'$ is
the set of linear orders that we started with to obtain $P_1$,
before removing some of the pairwise relations). Let $\vec
s_m=(m-1,\ldots,0)$. $P_2$ is a set of linear orders such that the
following holds for $Q=P_1'\cup
P_2\cup\{V\}$:\\
(1) For any $i\leq q$, $\vec s_m(Q,c)-\vec s_m(Q,v_i)=1$, $\vec
s_m(Q,w)-\vec
s_m(Q,c)=4q/3$.\\
(2) For any $i\leq q$, the scores of $v_i$ and $w,c$ are higher than
the score of $d$ in any extension of $P_1\cup P_2\cup\{V\}$ and in
any extension of $P_1\cup
P_2\cup\{U\}$.\\
(3) The size of $P_2$ is polynomial in $t+q$.

We now show how to construct $P_2$ in polynomial time. For any
alternative $a\neq d$, we define the following two votes:
$W_{a}=\{[a\succ d\succ\text{Others}], [\rev(\text{Others})\succ
a\succ d]\}$, where $\rev(\text{Others})$ is the reversed order of
the alternatives in $\mc\setminus\{a,d\}$. We note that for any
alternative $a'\in\mc\setminus \{a,d\}$, $\vec s_m(W,a)-\vec
s_m(W,a')=1$ and $\vec s_m(W,a')-\vec s_m(W,d)=1$. Let
$Q_1=P_1'\cup\{V\}$. $P_2$ is composed
of the following parts:\\
(1) $tm-\vec s_m(Q_1,c)$ copies of $W_{c}$.\\
(2) $tm+4q/3-\vec s_m(Q_1,w)$ copies of $W_{w}$.\\
(2) For each $i\leq q$, there are $tm-1-\vec s_m(Q_1,v_i)$ copies of
$W_{v_i}$.

We next prove that $V$ is dominated by $U$ if and only if $c$ is the
winner in at least one extension of $P_{po}\cup \{V\}$. We note that
for any $v\in \mv\cup\{w\}$, the score of $v$ in $V$ is the same as
the score of $v$ in $U$. The score of $c$ in $U$ is lower than the
score of $c$ in $V$. Therefore, for any extension $P^*$ of $P_{po}$,
if $r(P^*\cup\{V\})\in (\{w\}\cup \mv)$, then
$r(P^*\cup\{V\})=r(P^*\cup\{U\})$ (because $d$ cannot win). Hence,
for any extension $P^*$ of $P_{po}$, voting $U$ can result in a
different outcome than voting $V$ only if  $r(P^* \cup {V}) = c$. If
there exists an extension $P^*$ of $P_{po}$ such that
$r(P^*\cup\{V\})=c$, then we claim that the manipulator is strictly
better off voting $U$ than voting $V$. Let $P_1^*$ denote the
extension of $P_1$ in $P^*$. Then, because the total score of $w$ is
no more than the total score of $c$, $w$ is ranked lower than $d$ at
least $\frac{q}{3}$ times in $P_1^*$. Meanwhile, for each $i\leq q$,
$v_i$ is not ranked higher than $w$ more than one time in $P_1^*$,
because otherwise the total score of $v_i$ will be strictly higher
than the total score of $c$. That is, the votes in $P_1^*$ where
$d\succ w$ make up a solution to the {\sc x3c} instance. Therefore,
the only possibility for $c$ to win is for the scores of $c,w$, and
all alternatives in $\mv$ to be the same (so that $c$ wins according
to the tie-breaking mechanism). Now, we have $w=r(P^*\cup\{U\})$.
Because $w\succ_{V_M} c$, the manipulator is better off voting $U$.
It follows that $V$ is dominated by $U$ if and only if there exists
an extension of $P_{po}\cup \{V\}$ where $c$ is the winner.

The above reasoning also shows that $V$ is dominated by $U$ if and
only if the {\sc x3c} instance has a solution. Therefore, {\sc
domination} is {\sf NP}-hard. \myOmit{ For the {\sc dominating
manipulation} problem, we add to $P_{po}$ a profile $P_E$ defined as
follows. For each $e\in \mv\cup\{w\}$ and each $i\leq l-1$, we
obtain a vote $V_{e,i}$ from $V_M$ by exchanging the alternative
ranked in the $(i+1)$th position and $e$, and then exchanging the
alternative ranked in the $i$th position and $d$; let $O_{e,i}$
denote the partial order obtained from $V_{e,i}$ by removing $d\succ
e$. Let $M$ denote the following cyclic permutation $c\ra w\ra d\ra
\mv\ra A\ra c$. Let $P_E$ denote $q$ copies of
$\{O_{e,i},M(V_{e,i}),M(V_{e,i})^2,\ldots,M^{l-1}(V_{e,i}):e\in
\mv\cup\{w\}, i\leq l-1\}$. We note that in an extension $P_E^*$ of
$P_E$ where the extension of $O_{e,i}$ is $V_{e,i}$, then the scores
of the alternatives in $P_E^*$ are the same.

For any vote $W$ where there exists $v\in \mv$ such that the score
difference between $w$ and $v$ is different from the score
difference between $w$ and $v$ in $V_M$, there must exists $v'\in
\mv$ such that the score difference between $w$ and $v'$ in $W$ is
strictly smaller than their score difference in $V_M$. Then, it is
not hard to find an extension of $P_{po}$ such that if the
manipulator votes $V_M$, then $w$ wins, and if the manipulator votes
$W$, then $v'$ wins, which means that $V_M$ is not dominated by $W$.
Therefore, if $V_M$ is dominated by another $W$, then the score
differences between $w$ and the alternatives in $\mv$ are the same
across $V_M$ and $W$. Following the same reasoning as for the {\sc
domination} problem, we conclude that {\sc dominating manipulation}
is {\sf NP}-hard.}
\end{proof}

Theorem~\ref{thm:hardborda} can be generalized to a class of scoring
rules similar to the class of rules in Theorem~1
in~\cite{Xia11:Determining}, which does not include plurality or
veto. In fact, as we will show later, plurality and veto are
vulnerable to dominating manipulation.

We now investigate the relationship to the possible winner problem
in more depth. In a possible winner problem $(r,P_{po},c)$, we are
given a voting rule $r$, a profile $P_{po}$ composed of $n$ partial
orders, and an alternative $c$. We are asked whether there exists an
extension $P$ of $P_{po}$ such that $c=r(P)$. Intuitively, both {\sc
domination} and {\sc dominating manipulation} seem to be harder than
the possible winner problem under the same rule. Next, we present
two theorems, which show that for any WMG-based rule, {\sc
domination} and {\sc dominating manipulation} are harder than two
special possible winner problems, respectively.
%Therefore, if computing such restricted possible
%winner problems is {\sf NP}-hard, then {\sc domination} and {\sc
%dominating manipulation} are also {\sf NP}-hard.

We first define a notion that will be used in defining the two
special possible winner problems.
 For any instance of the possible winner
problem $(r,P_{po},c)$, we define its {\em WMG partition}
$\mr=\{R_{c'}:c'\in\mc \}$ as follows. For any $c'\in\mc$, let
$R_{c'}=\{\wmg(P):P\text{ extends }P_{po}\text{ and }r(P)=c'\}$.
That is, $R_{c'}$ is composed of all WMGs of the extensions of
$P_{po}$, where the winner is $c'$.
%\begin{itemize}
%\item $\bigcup_{i=1}^m R_i$ is the set of all weighted majority graphs corresponding to the extensions of
%$P_{po}$. That is, $\bigcup_{i=1}^m R_i=\{\wmg(P): P\text{ extends }
%P_{po}\}$.
%\item For any $i_1\neq i_2$, $R_{i_1}\cap R_{i_2}=\emptyset$, and
%for any $i\leq m$, any $G\in R_{i}$, the winner for $G$ is $c_i$.
%\end{itemize}
It is possible that for some $c'\in\mc$, $R_{c'}$ is empty. For any
subset $\mc'\subseteq \mc\setminus\{c\}$, we let $G_{\mc'}$ denote
the weighted majority graph where for each $c'\in \mc'$, there is an
edge $c'\ra c$ with weight $2$, and these are the only edges in
$G_{\mc'}$.
%In a {\em possible winner} problem, we are given $(r,P_{po},c)$,
%where $r$ is a voting rule, $P_{po}$ is a profile that is composed
%of $n$ partial orders, and $c$ is an alternative. We are asked
%whether there exists an extension $P$ of $P_{po}$ such that
%$c=r(P)$.
We are ready to define the two special possible winner problems for
WMG-based voting rules. \vspace{-1mm}
\begin{definition} Let $d^*$ be an alternative and let $\mc'$ be a nonempty subset of
$\mc\setminus\{c,d^*\}$. For any WMG-based voting rule $r$, we let
PW$_1(d^*,\mc')$ denote the set of possible winner problems $(r,
P_{po},c)$ satisfying the following conditions:
\begin{enumerate}
\item For any $G\in R_c$, $r(G+G_{\mc'})=d^*$.
\item For any $c'\neq c$ and any $G\in R_{c'}$,
$r(G+G_{\mc'})=r(G)$.
\item For any $c'\in \mc'$, $R_{c'}=\emptyset$.
\end{enumerate}
\end{definition}

We recall that $R_{c}$ and $R_{c'}$ are elements in the WMG
partition of the possible winner problem. \vspace{-1mm}
\begin{definition}Let $d^*$ be an alternative and let $\mc'$ be a nonempty subset of
$\mc\setminus\{c,d^*\}$. For any WMG-based voting rule $r$, we let
PW$_2(d^*,\mc')$ denote the problem instances $(r, P_{po},c)$ of
PW$_1(d^*,\mc')$, where for any $c'\in \mc\setminus\{c,d^*\}$,
$R_{c'}=\emptyset$.
\end{definition}\vspace{-2mm}
\begin{thm}
\label{thm:WMGdomhard}Let $r$ be a WMG-based voting rule. There is a
polynomial time reduction from PW$_1(d^*,\mc')$ to {\sc domination}
with partial orders, both under $r$.
\end{thm}\vspace{-1mm}
\begin{proof} Let $(r,P_{po},c)$ be a PW$_1(d^*,\mc')$
instance. We construct the following {\sc domination} instance. Let
the profile of partial orders be $Q_{po}=P_{po}\cup \{\rev(d^*\succ
c\succ \mc'\succ \others)\}$, $V=V_M=[d^*\succ c\succ \mc'\succ
\others]$, and $U=[d^*\succ \mc'\succ c\succ \others]$. Let $P$ be
an extension of $P_{po}$. It follows that $\wmg(P\cup\{\rev
(d^*\succ c\succ \mc'\succ \others), V\})=\wmg(P)$, and
$\wmg(P\cup\{\rev (d^*\succ c\succ \mc'\succ \others),
U\})=\wmg(P)+G_{\mc'}$. Therefore, the manipulator can change the
winner if and only if $\wmg(P)\in R_c$, which is equivalent to $c$
being a possible winner. We recall that by the definition of
PW$_1(d^*,\mc')$, for any $G\in R_c$, $r(G+G_{\mc'})=d^*$; for any
$c'\neq c$ and any $G\in R_{c'}$, $r(G+G_{\mc'})=c'$; and
$d^*\succ_V c$. It follows that $V$ (=$V_M$) is dominated by $U$ if
and only if the PW$_1(d^*,\mc')$ instance has a solution.
\end{proof}

Theorem~\ref{thm:WMGdomhard} can be used to prove that {\sc
domination} is {\sf NP}-hard for Copeland, maximin, and voting
trees, even when the number of undetermined pairs in each partial
order is bounded above by a constant. It suffices to show that for
each of these rules, there exist $d^*$ and $\mc'$ such that
PW$_1(d^*,\mc')$ is {\sf NP}-hard. To prove this, we can modify the
{\sf NP}-completeness proofs of the possible winner problems for
Copeland, maximin, and voting trees by Xia and
Conitzer~\cite{Xia11:Determining}. These proofs are omitted due to
space constraint.\vspace{-1mm}
\begin{corollary}
\label{coro:WMGhard} {\sc domination} with partial orders is {\sf
NP}-hard for Copeland, maximin, and voting trees, even when the
number of unknown pairs in each vote is bounded above by a constant.
\end{corollary}

\vspace{-2mm}
\begin{thm}
\label{thm:WMGDMhard}Let $r$ be a WMG-based voting rule. There is a
polynomial-time reduction from PW$_2(d^*,\mc')$ to {\sc dominating
manipulation}  with partial orders, both under $r$.
\end{thm}\vspace{-1mm}
\begin{proof} The proof is similar to the proof for
Theorem~\ref{thm:WMGdomhard}. We note that $d^*$ is the
manipulator's top-ranked alternative. Therefore, if $c$ is not a
possible winner, then $V$ ($=V_M$) is not %weakly
dominated by any
other vote; if $c$ is a possible winner, then $V$ is dominated by
$U=[w\succ \mc'\succ c\succ\others]$.\end{proof}

Similarly, we have the following corollary.
\begin{corollary}
\label{coro:WMGDMhard} {\sc dominating manipulation} with partial
orders is {\sf NP}-hard for Copeland and voting trees, even when the
number of unknown pairs in each vote is bounded above by a constant.
\end{corollary}
It is an open question if PW$_2(d^*,\mc')$  with partial orders is
{\sf NP}-hard for maximin. However, we can directly prove that {\sc
dominating manipulation} is {\sf NP}-hard for maximin by a reduction
from {\sc x3c}.
\begin{thm}
\label{thm:DMmaximin} {\sc dominating manipulation} with partial
orders is {\sf NP}-hard for maximin, even when the number of unknown
pairs in each vote is no more than $4$.
\end{thm}

For plurality and veto, there exist polynomial-time algorithms for
both {\sc domination} and {\sc dominating manipulation}. Given an
instance of {\sc domination}, denoted by $(r,P_{po}, V_M, V, U)$, we
say that $U$ is a {\em possible improvement} of $V$, if there exists
an extension $P$ of $P_{po}$ such that $r(P\cup\{U\})\succ_{V_M}
r(P\cup\{V\})$. It follows that $U$ dominates $V$ if and only if $U$
is a possible improvement of $V$, and $V$ is not a possible
improvement of $U$. We first introduce an algorithm
(Algorithm~\ref{alg:pi}) that checks whether $U$ is a possible
improvement of $V$ for plurality.

Let $c_{i^*}$ (resp., $c_{j^*}$) denote the top-ranked alternative
in $V$ (resp., $U$).
%The key idea
%behind Algorithm~\ref{alg:pi} is that we check for all pairs of
%alternatives $d,d'$ such that (1) $d'\succ_{V_M} d$, (2) $d$ wins if
%the manipulator votes for $V$, and (3) $d'$ wins if the manipulator
%votes for $U$. It follows that either $d=c_{i^*}$ or $d'=c_{j^*}$
%(or both hold).
We will check whether there exists $0\leq l\leq n$, $d,d'\in\mc$
with $d'\succ_{V_M}d$, and an extension $P^*$ of $P_{po}$, such that
if the manipulator votes for $V$, then the winner is $d$, whose
plurality score in $P^*$ is $l$, and if the manipulator votes for
$U$, then the winner is $d'$. We note that if such $d,d'$ exist,
then either $d=c_{i^*}$ or $d'=c_{j^*}$ (or both hold). To this end,
we solve multiple maximum-flow problems defined as follows.

Let $\mc'\subset \mc$ denote a set of alternatives. Let $\vec
e=(e_1,\ldots,e_m)\in {\mathbb N}^m$ be an arbitrary vector composed
of $m$ natural numbers such that $\sum_{i=1}^m e_i\geq n$. We define
a maximum-flow problem $F_{\mc'}^{\vec e}$ as follows.\\
{\bf Vertices:} $\{s, O_1,\ldots,O_n, c_1,\ldots,c_m, y,t\}$.\\
{\bf Edges:}
\begin{itemize}
\item For any $O_i$, there is an edge from $s$ to $O_i$ with capacity $1$.
\item For any $O_i$ and $c_j$, there is an edge $O_i\ra c_j$ with
capacity $1$ if and only if $c_j$ can be ranked in the top position
in at least one extension of $O_i$.
\item For any $c_i\in \mc'$, there is an edge $c_i\ra t$ with capacity
$e_i$.
\item For any $c_i\in \mc\setminus \mc'$, there is an edge $c_i\ra y$ with capacity
$e_i$.
\item There is an edge $y\ra t$ with capacity
$n-\sum_{c_i\in\mc'}e_i$.
\end{itemize}

\begin{figure}[h]
\centering
\includegraphics{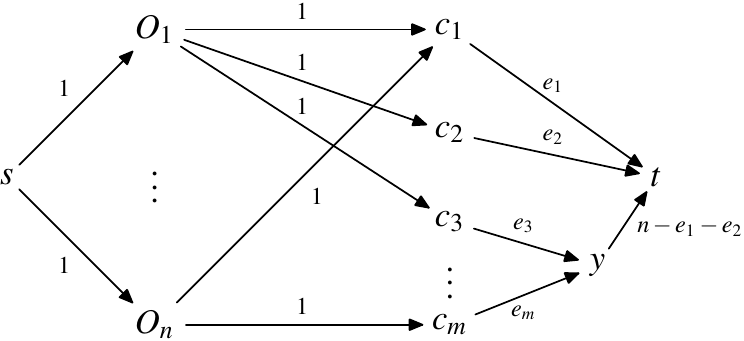}\vspace{-2mm}
\caption{\footnotesize$F_{\{c_1,c_2\}}^{\vec e}$.}%\vspace{-2mm}
\label{ex:fdomination}
\end{figure}

For example, $F_{\{c_1,c_2\}}^{\vec e}$ is illustrated in
Figure~\ref{ex:fdomination}.

It is not hard to see that $F_{\mc'}^{\vec e}$ has a solution whose
value is $n$ if and only if there exists an extension $P^*$ of
$P_{po}$, such that (1) for each $c_i\in \mc'$, the plurality of
$c_i$ is exactly $e_i$, and (2) for each $c_{i'}\not\in \mc'$, the
plurality of $c_{i'}$ is no more than $e_{i'}$. Now, for any pair of
alternatives $d=c_{i},d'=c_{j}$ such that $d'\succ_{V_M} d$ and
either $d=c_{i^*}$ or $d'=c_{j^*}$, we define the set of {\em
admissible maximum-flow problems} $A^l_{\text{Plu}}$ to be the set
of maximum flow problems $F_{c_i,c_j}^{\vec e}$ where $e_i=l$, and
if $F_{c_i,c_j}^{\vec e}$ has a solution, then the manipulator can
improve the winner by voting for $U$. Details are omitted due to
space constraint. Algorithm~\ref{alg:pi} solves all maximum-flow
problems in $A^l_{\text{Plu}}$ to check whether $U$ is a possible
improvement of $V$.\vspace{-3mm}
\begin{algorithm}
%\SetLine
\caption{PossibleImprovement($V$,$U$)\label{alg:pi}}\DontPrintSemicolon
Let $c_{i^*}=\alt(V,1)$ and $c_{j^*}=\alt(U,1)$.\;

\For {any $0\leq l\leq n$ and any pair of alternatives
$d=c_{i},d'=c_{j}$ such that $d'\succ_{V_M} d$ and either
$d=c_{i^*}$ or $d'=c_{j^*}$}{

Compute $A^l_{\text{Plu}}$.\;

\For { each maximum-flow problem $F_{\mc'}^{\vec e}$ in
$A^l_{\text{Plu}}$} {

\If{$\sum_{c_i\in \mc'} e_i\leq n$ and the value of maximum flow in
$F_{\mc'}^{\vec e}$ is $n$}{ Output that the $U$ is a possible
improvement of $V$, terminate the algorithm. }
 }
} Output that $U$ is not a possible improvement of $V$.\;
\end{algorithm}

\vspace{-4mm}
 The algorithm for {\sc domination}
(Algorithm~\ref{alg:domination}) runs Algorithm~\ref{alg:pi} twice
to check whether $U$ is a possible improvement of $V$, and whether
$V$ is a possible improvement of $U$.
\begin{algorithm}[h]
\SetAlgoSkip{} %\SetLine
\caption{Domination\label{alg:domination}}\DontPrintSemicolon

\If{PossibleImprovement($V$,$U$)=``yes'' and
PossibleImprovement($U$,$V$)=``no''}{

Output that $V$ is dominated by $U$.}\Else{Output that $V$ is not
dominated by $U$.}
\end{algorithm}
The algorithm for {\sc dominating manipulation} for plurality simply
runs Algorithm~\ref{alg:domination} $m-1$ times. In the input we
always have that $V=V_M$, and for each alternative in
$\mc\setminus\{\alt(V,1)\}$, we solve an instance where that
alternative is ranked first in $U$. If in any step $V$ is dominated
by $U$, then there is a dominating manipulation; otherwise $V$ is
not dominated by any other vote. The algorithms for {\sc domination}
and {\sc dominating manipulation} for veto are similar. We omit the
details due to space constraint. \vspace{-2mm}
\section{Future Work}
Analysis of manipulation with partial information provides insight
into what needs to be kept confidential in an election. For
instance, in a plurality or veto election, revealing (perhaps
unintentionally) part of the preferences of non-manipulators may
open the door to strategic voting. An interesting open question is
whether there are any more general relationships between the
possible winner problem and the dominating manipulation problem with
partial orders. It would be interesting to identify cases where
voting rules are resistant or even immune to manipulation based on
other types of partial information, for example, the set of possible
winners. We may also consider other types of strategic behavior with
partial information in our framework, for example, coalitional
manipulation, bribery, and control. We are currently working on
proving completeness results for higher levels of the polynomial
hierarchy for problems similar to those studied in this paper.
%Finally, we could
%also perform game theoretic analysis of sequential voting games
%where participants only have partial information about the current
%state.
%\bibliography{refshort}
%\bibliographystyle{plain}
%\bibliographystyle{aaai}
\vspace{-2mm}
\section*{Acknowledgments}\vspace{-1mm}
Vincent Conitzer and Lirong Xia acknowledge NSF CAREER 0953756 and
IIS-0812113, and an Alfred P.~Sloan fellowship for support. Toby
Walsh is supported by the Australian Department of Broadband,
Communications and the Digital Economy, the ARC, and the Asian
Office of Aerospace Research and Development (AOARD-104123). Lirong
Xia is supported by a James B.~Duke Fellowship. We thank all AAAI-11
reviewers for their helpful comments and suggestions. 

\end{document}